\documentclass[11pt]{article}

\usepackage[colormark, titleright, nolineno]{showlab}
\shorttitle{Show-Harness}

\usepackage{tabularx}
\usepackage{colortbl}   
\usepackage{nicematrix}
\usepackage{subcaption}   
\usepackage{wrapfig}     

\definecolor{slGray}{HTML}{6B7684}   
\definecolor{slInk}{HTML}{2C3440}   
\newcommand{\MODEL}{\textcolor{slBlue}{S}\textcolor{slGreen}{h}\textcolor{slRed}{o}\textcolor{slYellow}{w}-Harness}

\logo{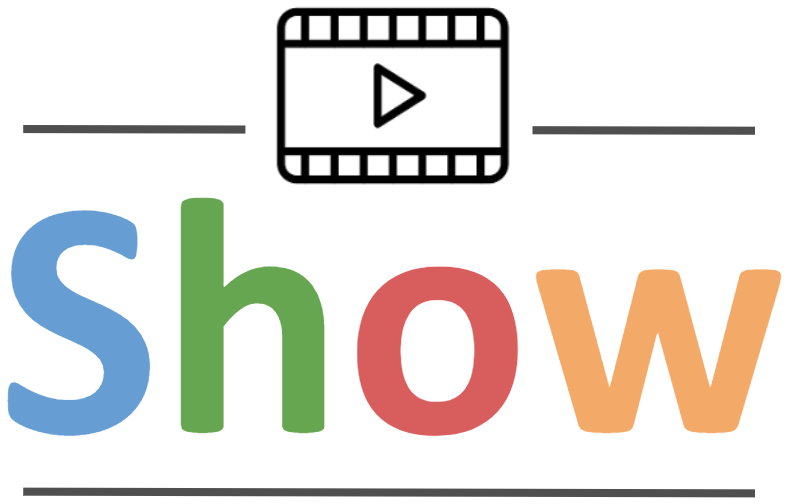}   
\logoheight{16mm}       
\headlogoheight{6.5mm}  
\runningmark{September, 2026} 

\newcommand{\Rom}[1]{\uppercase\expandafter{\romannumeral #1}}

\title{\MODEL: Just a VLM Agent Can Play Robots}

\author{%
  Yanzhe Chen\amark[*]{1}\quad
  Zechen Bai\amark[*]{1}\quad
  Zhijun Cao\amark[*]{1}\quad
  Wenzheng Zeng\amark[*]{1}\quad
  Kevin Qinghong Lin\quad \\
  Yiqi Lin\amark{1}\quad
  Guoqiang Liang\amark{1}\quad
  Kevin Yuchen Ma\amark{1}\quad
  Qiming Huang\amark{1}\quad
  Mike Zheng Shou\amark[\textdagger]{1}%
}
\affiliation{Show Lab, National University of Singapore}
\authornote{\textsuperscript{*}Equal contribution.\quad
            \textsuperscript{\textdagger}Corresponding author}
\links{%
  \resource{\textbf{Website [Code \& Model \& Dataset]} :}{https://showlab.github.io/Show-Harness}%
}

\begin{document}
\maketitle
\thispagestyle{firstpage}

\begin{abstract}
Foundation vision--language models (VLMs) exhibit broad intelligence about the world, yet translating this intelligence into robot control remains challenging.
We present \textbf{Show-Harness}, an \emph{Embodied Harness} that enables VLMs to \textbf{``play'' robots} through a compact semantic interface linking intent to action.
Show-Harness exposes discrete semantic action units that VLMs can naturally reason over, while embodiment-specific interpreters deterministically ground them into local robot actions, keeping the VLM directly responsible for fine-grained physical decisions.
Through the same interface, Show-Harness demonstrates the feasibility of (1) directly unlocking closed-source frontier VLMs for zero-shot robot control, and (2) adapting small-scale open-source VLMs for low-cost deployment with just a few GPU-hours of fine-tuning.
We further develop \textbf{GUMI} (\textbf{GU}I \textbf{M}anipulation \textbf{I}nterface), which extends the same semantic action space to GUI-based demonstration collection, allowing humans and agents to ``play'' robots across embodiments without specialized teleoperation hardware.
Extensive experiments show that Show-Harness-equipped VLM agents generalize robustly across tasks, embodiments, and environments, outperforming representative agentic and VLA paradigms.
These results suggest that the right interface can unlock substantial embodied capability from foundation VLMs, without requiring additional model capacity or costly embodiment-specific pretraining.

\end{abstract}



\begin{figure}[h]
  \centering
  \includegraphics[width=\linewidth]{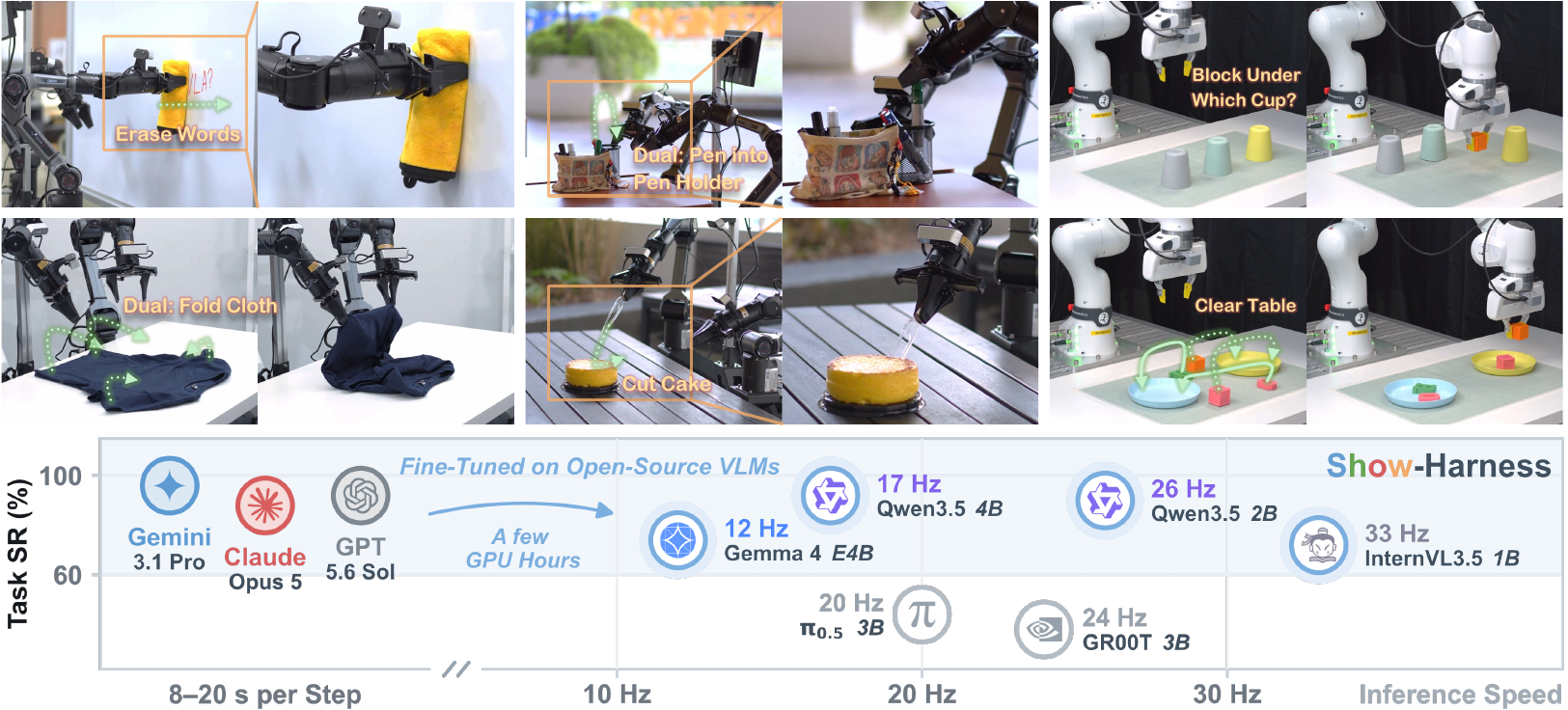}
   \caption{Show-Harness unlocks generalizable embodied manipulation with foundation VLMs across diverse tasks, scenes, and robot embodiments, enabling direct zero-shot deployment with frontier models and capable, efficient control via lightweight adaptation of compact open-source models. 
   }
  \label{fig:topdemo}
\end{figure}

\begin{figure}[t]
  \centering
  \includegraphics[width=\linewidth]{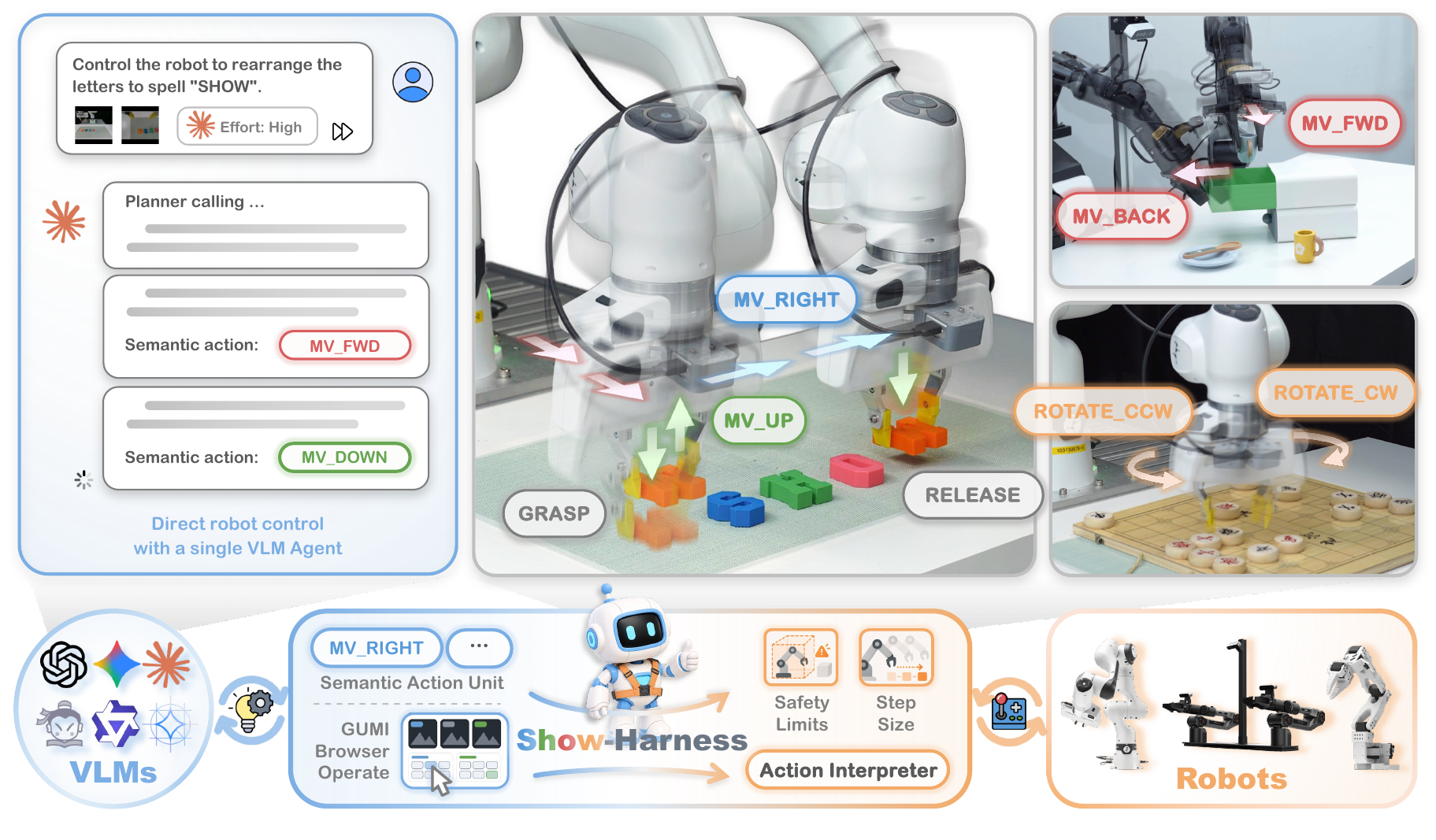}
  \vspace{-8mm}
  \caption{Show-Harness connects foundation VLMs to diverse robot embodiments through one
semantic action interface, enabling zero-shot control with frontier models and
lightweight adaptation of small open models.}
  \label{fig:qiancai}
\end{figure}

\section{Introduction}
\label{sec:intro}

Foundation vision--language models (VLMs) already encode much of what robot manipulation requires, from recognizing objects and spatial relations to decomposing long-horizon goals~\citep{driess2023palm,yuan2024robopoint,ji2025robobrain,team2025gemini}.
Yet this knowledge does not readily translate into robot behavior.
Vision--language--action (VLA) models pull VLMs toward low-level control by fine-tuning them to regress embodiment-specific continuous actions, collapsing broad semantic knowledge into an opaque pixel-to-actuation mapping that often requires repeated adaptation across tasks, environments, and embodiments~\citep{brohan2023rt,kim2024openvla,black2024pi_0,o2024open}.
Hierarchical and programmatic systems take the opposite route: the VLM produces explicit intermediate abstractions, including subtask-level calls~\citep{ahn2022can,huang2022inner,shi2025hi,yang2025agentic} and programs over hand-designed control APIs~\citep{liang2023code,huang2023instruct2act,chen2024roboscript}. Despite the effectiveness, the physical control is mediated by downstream controllers or system-specific mechanisms, weakening the direct link between semantic intent and physical execution.

\textbf{We argue that bringing foundation-model intelligence into the physical world requires a suitable interface:}
an action space that is semantically meaningful to the VLM yet sufficiently fine-grained for direct physical control.
We introduce \textbf{Show-Harness}, a model-agnostic \emph{Embodied Harness} that enables VLMs to \textbf{``play'' robots} through such an interface.
As shown in Fig.~\ref{fig:qiancai}, Show-Harness reformulates robot control into a discrete set of semantic action units. Each unit specifies a single movement step in a given direction or a gripper action, making it directly interpretable to the VLM. An embodiment-specific interpreter then deterministically grounds each unit into a small, bounded robot motion, letting every semantic decision land in the physical world.
Around this core, the harness closes the loop: it organizes multi-view observations and proprioception into a perceptual context, supports subtask reasoning and recovery, and returns execution feedback after every unit, turning digital VLMs into \emph{situated agents} that reason and act in a semantic space they natively understand while remaining directly responsible for fine-grained physical decisions.

Show-Harness enables two modes of robot control. First, it seamlessly turns closed-source frontier VLMs into zero-shot robot agents without fine-tuning, outperforming representative agentic harnesses across tasks, embodiments, and environments. 
Second, it enables lightweight VLMs (e.g., 2B-scale models) to ``play'' robots in the same action space with just a few GPU-hours of fine-tuning, achieving stronger generalization and sim-to-real transfer than representative VLA paradigms under controlled experiments.
Together, these results suggest that a suitable semantic interface can unlock substantial embodied capability from foundation VLMs, providing a scalable path that inherits advances in frontier models while extending such control to smaller, lower-cost models.

Since the action space is interpretable and directly operable by both humans and models, Show-Harness naturally unifies VLM-based robot agents with GUI-style human control.
We further build \textbf{GUMI}, a \textbf{GU}I-based \textbf{M}anipulation \textbf{I}nterface that enables both humans and agents to collect robot demonstrations in the same semantic action space, without specialized teleoperation hardware, while naturally supporting cross-embodiment reuse and human--agent collaborative data collection.
The main contributions of this work are summarized as follows:
\begin{itemize}

\item We introduce \textbf{Show-Harness}, a model-agnostic \emph{Embodied Harness} that enables foundation VLMs to directly operate robots through a compact semantic action interface.

\item Show-Harness demonstrates the feasibility of directly unlocking closed-source frontier VLMs for zero-shot control, and efficiently adapting small-scale models for low-cost deployment.

\item Show-Harness-equipped VLM agents demonstrate strong generalization across tasks, embodiments, and environments, outperforming representative agentic and VLA paradigms. Further studies on physical and semantic adaptability reveal key properties of effective VLM--robot interfaces.

\item We introduce \textbf{GUMI}, a GUI-based manipulation interface that enables humans, agents, and human--agent collaboration to collect robot demonstrations in the same semantic action space across embodiments, without specialized teleoperation hardware.
\end{itemize}

\section{Related Work}
\label{sec:related_work}

\subsection{Foundation Models for Robot Manipulation}

\noindent \textbf{Foundation models as low-level policies.}
Vision--language--action (VLA) models attach learned action generation to pretrained VLM backbones and predict low-level robot controls. These may take the form of continuous action chunks produced by regression, diffusion, or flow matching~\citep{zhao2023learning,chi2025diffusion,black2024pi_0,liu2025rdt,shukor2025smolvla,team2025gemini,hu2026ar}; discretized motor tokens~\citep{brohan2022rt,brohan2023rt,kim2024openvla,pertsch2025fast,jiang2026sa}; keyframe or spatial-grid actions in task space~\citep{shridhar2023perceiver,qu2025spatialvla}; or latent action codes learned from robot trajectories and videos~\citep{lee2024behavior,ye2025latent,chen2025moto,zheng2025universal,bu2025univla,bauer2025latent,mu2026one,kang2026x}.
Recent extensions couple action prediction with future visual dynamics in unified video--action or world-action models~\citep{li2025unified,zhu2025unified,ye2026world,li2026light,pan2026selfwam,yang2026lila}. Robot foundation models further scale learned action generation across heterogeneous tasks and embodiments~\citep{o2024open,bousmalis2023robocat,team2024octo,doshi2024scaling,bjorck2025gr00t,generalist2026gen1,florence2026going,black2025pi_05}, while follow-up work improves action initialization, tokenization, and adaptation to new embodiments and tasks~\citep{goyal2025vla,xu2026apt,jing2026learning,wang2026context,chen2026escaping,bhatia2026adapting}.
This route, however, trades semantics for control. Re-fitting the model to embodiment-specific motor signals collapses its broad pretrained knowledge into an opaque sensorimotor mapping and usually requires fresh robot trajectories for new task families or embodiments~\citep{zhong2025survey,karcini2026robots}.

\noindent \textbf{Foundation models as intermediate decision makers.}
Another route keeps the VLM above low-level control, using pretrained semantic knowledge to decompose instructions and emit subgoals, keypoints, affordance targets, value maps, or spatial constraints~\citep{driess2023palm,yuan2024robopoint,ji2025robobrain,huang2023voxposer,duan2024manipulate,huang2024rekep}. Downstream controllers then handle their physical realization.
This preserves semantics but surrenders physics: the model specifies intent without seeing how it is realized~\citep{hu2026matters,galanti2026addressing}, and each representation is bound to a carefully engineered, system-specific grounding pipeline~\citep{huang2023voxposer,duan2024manipulate,huang2024rekep,liu2024ok}.
Realistic deployment further requires long-horizon planning, closed-loop feedback, and failure recovery~\citep{galanti2026addressing}, pushing these hierarchical designs toward the agentic systems discussed next.

\subsection{Agentic Robot Systems}

\noindent \textbf{Agentic architectures.}
Agentic systems keep the foundation VLM intact within a harness that translates its decisions into robot behavior and feeds back outcomes, closing the perception--decision--execution loop~\citep{huang2022inner,yang2025agentic,liu2026guava,lima2026agentic}.
Systems differ mainly in how decisions are executed: the model may compose programs over perception and control APIs~\citep{liang2023code,singh2022progprompt,huang2023instruct2act,vemprala2024chatgpt,chen2024roboscript,fu2026cap,zhao2026rosclaw}, select skills from predefined libraries~\citep{ahn2022can,rana2023sayplan,ahn2024autort,yuan2025being,li2026roboclaw,santos2026alrm}, steer learned policies (e.g., VLAs) with language subgoals~\citep{shi2025hi,hu2026matters,zhang2026harness,chen2026volo,peng2026cortex}, or offload decisions to symbolic planners and numerical optimizers~\citep{liu2023llm+,chen2024autotamp,chen2025code,yu2023language}.
The harness also provides the machinery that sustains long-horizon interaction: task context and persistent memory~\citep{rana2023sayplan,zhang2026harness,anwar2025remembr}, feedback-driven monitoring and reflection~\citep{huang2022inner,shinn2023reflexion}, and failure detection for replanning and recovery~\citep{liu2023reflect,duan2024aha,yang2025agentic}.

\noindent \textbf{Interfaces for embodied execution.}
Across these paradigms, the VLM ultimately acts through primitives exposed by the surrounding system, making interface design a central question.
Frontier VLMs perform very differently under different levels of abstraction~\citep{berman2026claude,hu2026matters}: human-designed abstractions such as \texttt{place(object, target)} improve reliability, whereas composing low-level perception and control APIs remains difficult~\citep{fu2026cap,liu2026guava}.
The prevailing solution is therefore to expose entire skills, controllers, or VLAs as callable primitives~\citep{ahn2022can,shi2025hi,zhang2026harness,chen2026volo,peng2026cortex}: the agent decides \emph{what} to do, while an opaque executor determines \emph{how}~\citep{galanti2026addressing}.
Show-Harness instead exposes the \emph{how} itself through fine-grained semantic action units whose physical realization is deterministic and transparent, allowing the VLM to remain directly responsible for fine-grained physical decisions.

\noindent \textbf{From digital interfaces to physical manipulation.}
Show-Harness's discrete semantic action space echoes a broader interface principle in digital and simulated agents: exposing compact, interpretable actions operable by both humans and foundation models. Computer-use and game agents act through mouse, keyboard, or controller inputs~\citep{lin2025showui,qin2025ui,wang2025game,ouyang2026gameworld,zhang2026towards}, while simulated embodied agents use discrete or skill-level actions~\citep{li2023behavior,li2024embodied,yang2025embodiedbench,li2026humanclaw}.
In navigation, discrete directional primitives are native to simulators and benchmarks~\citep{savva2019habitat,krantz2020beyond}, and recent general-purpose agents can drive them competitively~\citep{zhou2026embodied}.  
Real-world manipulation, however, lacks a comparably simple and broadly operable interface: control is typically embodiment-specific, high-dimensional, and requires fine-grained spatial precision.
Show-Harness bridges this gap by deterministically grounding fine-grained semantic actions into physical robot motion, while exposing the same action space to both VLM agents and humans.
Our experiments further demonstrate its strong sim-to-real transfer capability.

\begin{figure}[t]
  \centering
  \includegraphics[width=\linewidth]{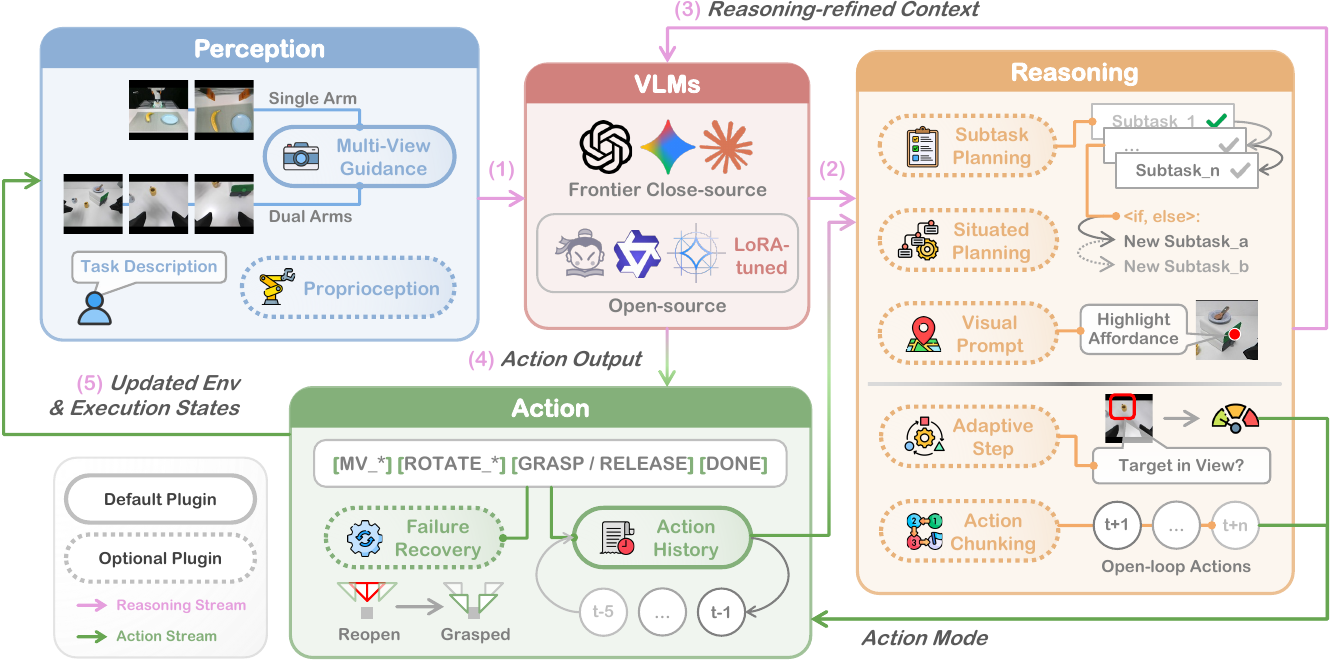}
  \caption{\textbf{The Show-Harness architecture.} A modular perceive--reason--act loop connects foundation VLMs to robot control through a shared semantic action interface.}
  \label{fig:method}
\end{figure}

\section{\MODEL}
\label{sec:method}

\newcommand{\plugin}[1]{\textbf{\texttt{#1}}}
\newcommand{\pon}{\textcolor{slGreen}{\checkmark}}
\newcommand{\popt}{\textcolor{slYellow}{$\boldsymbol{\circ}$}}
\newcommand{\poff}{\textcolor{slRed}{$\times$}}
\newcommand{\stageP}[1]{\textcolor{slBlue}{#1}}
\newcommand{\stageR}[1]{\textcolor{slYellow}{#1}}
\newcommand{\stageA}[1]{\textcolor{slGreen}{#1}}

\subsection{Overview}
\label{sec:setup}

Show-Harness is an embodied harness designed to translate the intelligence of foundation vision--language models into physically grounded robot behavior by placing the VLM inside an iterative perception--reasoning--action loop. At each interaction step, the model receives perceptual inputs and interaction history, reasons about what to do next, expresses its intent as a semantic action decision, and observes the resulting changes after that decision is grounded into physical robot motion.

Specifically, as shown in Fig.~\ref{fig:method}, given a language instruction $\ell$, at step $t$ the system captures an observation $o_t = \left(\mathcal{I}_t, p_t\right)$, where $\mathcal{I}_t$ denotes the multi-view visual observation and $p_t$ the robot's proprioceptive state.
The system first processes the current observation and a compact interaction history $h_t$ through a configurable set of reasoning plugins $\mathcal{P}$, producing a reasoning-refined context
\begin{equation}
  c_t = \Phi_{\mathcal{P}}\!\left(\ell,\, o_t,\, h_t\right).
  \label{eq:context}
\end{equation}
The central VLM then selects a semantic action conditioned on the refined context $c_t$,
\begin{equation}
  a_t = \pi(c_t) \in \mathcal{A},
  \label{eq:decision}
\end{equation}
where $\mathcal{A}$ is a compact set of fine-grained, actionable units that defines the semantic interface between the VLM and the robot. Each unit specifies a directly interpretable end-effector movement or gripper intent without exposing embodiment-specific control variables to the model (Sec.~\ref{sec:interface}).
The resulting semantic decision $a_t$ is then passed to an embodiment-specific and model-agnostic interpreter, which deterministically grounds it into executable robot control,
\begin{equation}
  u_t = g_E(a_t;\, s_t),
  \label{eq:ground}
\end{equation}
where $E$ denotes the robot embodiment and $s_t$ is the interpreter's internal setpoint state. Executing $u_t$ updates the robot and environment, producing new observations and execution states that are fed back into the next interaction step.

\subsection{Physically Grounded Semantic Action Interface}
\label{sec:interface}

At the core of Show-Harness is an interface that keeps VLM decisions semantically meaningful yet sufficiently fine-grained for direct physical control.

\paragraph{Semantic action space}
The shared action space $\mathcal{A}$ in Eq.~\ref{eq:decision} consists of a compact set of semantic action units.
At each step, the system determines a reference view from the current observations to define end-effector movement directions.
Relative to this view, \texttt{MV\_FWD}/\texttt{MV\_BACK}, \texttt{MV\_LEFT}/\texttt{MV\_RIGHT}, and \texttt{MV\_UP}/\texttt{MV\_DOWN} move the end effector one step along the corresponding direction. 
For tasks that require end-effector reorientation, \texttt{ROTATE\_CW} and \texttt{ROTATE\_CCW}, each paired with a specified axis (\(x\), \(y\), or \(z\)), incrementally rotate the end effector clockwise or counterclockwise about that axis.
\texttt{GRASP} and \texttt{RELEASE} close and open the gripper, while \texttt{DONE} indicates task completion.

The vocabulary is designed around 3 properties.
(i) \emph{Incremental}: each unit induces a small, localized physical change, keeping the VLM situated in the control loop through observable action effects and enabling precise behavior through successive corrections. We empirically find that VLMs can flexibly switch between step granularities without fine-tuning (Sec.~\ref{sec:analysis}).
(ii) \emph{Interpretable and embodiment-agnostic}: actions are represented as semantic symbols rather than numeric targets, while embodiment-specific low-level control is delegated to the downstream interpreter, keeping the model-facing space compact, interpretable, and reusable across robots.
(iii) \emph{Visually grounded}: movement directions are defined relative to observable views, allowing spatial reasoning to map onto actions.

\paragraph{Embodiment grounding}
Each semantic decision $a_t \in \mathcal{A}$ made by the VLM is then passed to an embodiment-specific interpreter $g_E$, which deterministically grounds the unit $a_t$ into robot control. For an action unit, the interpreter updates the 6-DoF Cartesian pose setpoint $s_t=(\mathbf{x}_t,Q_t)$ as
\begin{equation}
  s_{t+1} = \Pi_E\!\left(
    \mathbf{x}_t + \sigma_t R_E d_a,\;
    \exp\!\left(\theta_t [R_E r_a]_{\times}\right) Q_t
  \right),
  \label{eq:interp}
\end{equation}
where $\mathbf{x}_t\in\mathbb{R}^3$ and $Q_t\in\mathrm{SO}(3)$ denote position and orientation. $d_a$ and $r_a$ encode translation and rotation, respectively: $d_a=0$ for rotation, while $r_a=0$ for translation and otherwise $r_a\in\{\pm\mathbf{e}_x,\pm\mathbf{e}_y,\pm\mathbf{e}_z\}$. Here $[\cdot]_{\times}$ is the skew-symmetric matrix operator. The calibrated increments $\sigma_t$ and $\theta_t$ set the translation and rotation magnitudes, while $R_E$ maps the semantic directions and axes into the motion frame of embodiment $E$. The projection $\Pi_E$ enforces embodiment-specific workspace and per-step translation or rotation limits. Gripper units bypass the pose update and map directly to open or close commands.
Different embodiments realize the same semantic units through different low-level controllers. For example, Franka tracks Cartesian setpoints with impedance control, AgileX uses inverse kinematics and streamed joint targets, and the simulator executes corresponding operational-space commands.
By isolating embodiment-specific control inside $g_E$, the semantic model interface remains unchanged across robots. Adapting to a new embodiment therefore requires only a new interpreter. Safety bounds, such as workspace and table-height limits, can be flexibly configured in the interpreter, and actions that violate them are blocked before execution to ensure safe physical interaction.

\subsection{Embodied Harness Architecture}
\label{sec:harness}

The Embodied Harness organizes the foundation model's interaction loop into three configurable stages: \emph{Perception}, \emph{Reasoning}, and \emph{Action}. 
Table~\ref{tab:plugins} summarizes the plugins instantiated at each stage.

\begin{table}[t]
  \centering
  \caption{Harness plugins organized along the perceive--reason--act loop.}
  \label{tab:plugins}
  \small
  \begin{tabularx}{\linewidth}{l X}
    \toprule
   \textbf{ Plugin} & \textbf{Function} \\
    \midrule
    \rowcolor{slBlue!15}
    \multicolumn{2}{l}{\emph{Perception}} \\
    \rowcolor{slBlue!5}
    \plugin{Multi-View Guidance}       & Guides the VLM on the roles and appropriate use of different camera views             \\
    \rowcolor{slBlue!5}
    \plugin{Proprioception}   & Translates robot state, contact, and gripper status into textual feedback        \\
    \addlinespace
    \rowcolor{slYellow!15}
    \multicolumn{2}{l}{\emph{Reasoning}} \\
    \rowcolor{slYellow!5}
    \plugin{Subtask Planning}     & Maintains an ordered subtask plan with visually checkable completion       \\
    \rowcolor{slYellow!5}
    \plugin{Situated Planning}        & Defers uncertain decisions and selectively replans as execution unfolds \\
    \rowcolor{slYellow!5}
    \plugin{Action Chunking}     & Adaptively chunks actions to balance control precision and execution efficiency   \\
    \rowcolor{slYellow!5}
    \plugin{Adaptive Step}    & Adaptively adjusts step size to balance precision and efficiency \\
    \rowcolor{slYellow!5}
    \plugin{Visual Prompt}    & Highlights and verifies task-relevant visual targets \\
    
    \addlinespace
    \rowcolor{slGreen!15}
    \multicolumn{2}{l}{\emph{Action}} \\
    \rowcolor{slGreen!5}
    \plugin{Action History}   & Summarizes recent actions and discourages oscillatory behavior             \\
    \rowcolor{slGreen!5}
    \plugin{Failure Recovery} & Detects execution failures and triggers corrective recovery             \\
    \bottomrule
  \end{tabularx}
\end{table}

\paragraph{\stageP{Perception}}
Perception plugins turn raw sensor streams into a context the model can reason over:
\begin{itemize}

  \item \textbf{Multi-View Guidance} tells the VLM how different camera views should be used and prioritized. For example, a global egocentric/exocentric view provides scene-level context, while a wrist-mounted view offers close-up visual evidence for fine-grained alignment and manipulation.

    \item \textbf{Proprioception} translates the robot's internal state into concise textual feedback, including the current gripper height, the displacement induced by one action step, phase-aware execution hints (e.g., descending first when still high), contact state, and gripper state.
\end{itemize}

\paragraph{\stageR{Reasoning}}
Reasoning plugins structure the task into actionable intermediate decisions and refine the current context for fine-grained action-level decisions:
\begin{itemize}
  \item \textbf{Subtask Planning} first invokes the VLM as a planner to decompose the task instruction $\ell$ into an ordered sequence of subtasks with completion criteria. During execution, subtask transitions belong to the model: at every step it checks the completion criterion against the current images, keeps acting while it is unmet, and advances the plan by declaring the subtask complete.

\item \textbf{Situated Planning} defers uncertain decisions and selectively triggers replanning when the required information becomes available during execution.
Rather than committing to all future branches upfront, the initial plan leaves such decisions unresolved until the relevant evidence becomes observable.
The VLM then resolves the branch from the current observation and updates the remaining plan accordingly.
This enables conditional tasks to adapt to the evolving execution state without replanning at every interaction step.

  \item \textbf{Action Chunking} adaptively reduces model-query frequency when fine-grained feedback is not yet required (e.g., when the target is still far away), allowing the VLM to emit a short sequence of semantic actions that are executed open-loop before the next model query.
  
 \item \textbf{Adaptive Step} dynamically adjusts the step size to balance efficiency and precision, using larger steps when the target is distant and smaller steps for close-range alignment.
  \item \textbf{Visual Prompt} uses a dedicated model call to convert an ambiguous verbal target into a visual reference highlighting the relevant affordance, allowing subsequent reasoning to ground on the visual marker. This plugin is enabled when the interaction region is difficult to specify in language.
  
\end{itemize}

\paragraph{\stageA{Action}}
Building on the physically grounded semantic action interface introduced in Sec.~\ref{sec:interface}, the action stage augments execution with lightweight interaction memory and recovery mechanisms that support robust closed-loop control.
Two plugins operate at this stage:
\begin{itemize}
  \item \textbf{Action History} carries recent actions into the next interaction step as lightweight context, together with simple usage guidance such as avoiding oscillation between opposite moves. This compact history provides the model with the temporal memory needed across steps.
  \item \textbf{Failure Recovery} automatically detects grasp failures and recovers by resetting the gripper state and rolling back to the relevant grasp subtask.

\end{itemize}

\subsection{Two Modes on One Interface}
\label{sec:modes}

The shared semantic interface supports two complementary modes of robot control.

\paragraph{Frontier VLM as zero-shot agents (ZS mode)}
A frontier VLM can directly control robots through the harness without any fine-tuning.
This mode directly unlocks frontier-model capabilities for physical control, providing a scalable path that inherits advances in increasingly capable foundation models.

\paragraph{Fine-tuning small VLMs (FT mode)}
The same interface supports lightweight adaptation of a small open-source VLM to predict semantic action units directly. Given demonstrations $\mathcal{D}$ collected in the shared action space, the policy minimizes the token-level cross-entropy of the target unit,
\begin{equation}
  \mathcal{L}(\theta) = -\!\!\sum_{(\ell,\, o,\, h,\, a)\, \in\, \mathcal{D}}\!\! \log \pi_\theta\!\left(a \mid \Phi_{\mathcal{P}_{\mathrm{min}}}(\ell,\, o,\, h)\right),
  \label{eq:sft}
\end{equation}
where $\mathcal{P}_{\mathrm{min}}$ intentionally retains only a minimal decision context to facilitate controlled comparison and analysis, consisting of the task instruction $\ell$, multi-view observation $o$, and a short action history $h$.
Crucially, semantic actions are predicted through the VLM's native vocabulary, without dedicated action heads or special tokens, enabling lightweight low-rank adaptation from limited demonstrations and stronger generalization than representative VLA baselines, as demonstrated in Sec.~\ref{sec:experiments}.

\section{GUMI: A GUI-based Manipulation Interface}
\label{sec:gumi}

\begin{figure}[t]
  \centering
  \includegraphics[width=\linewidth]{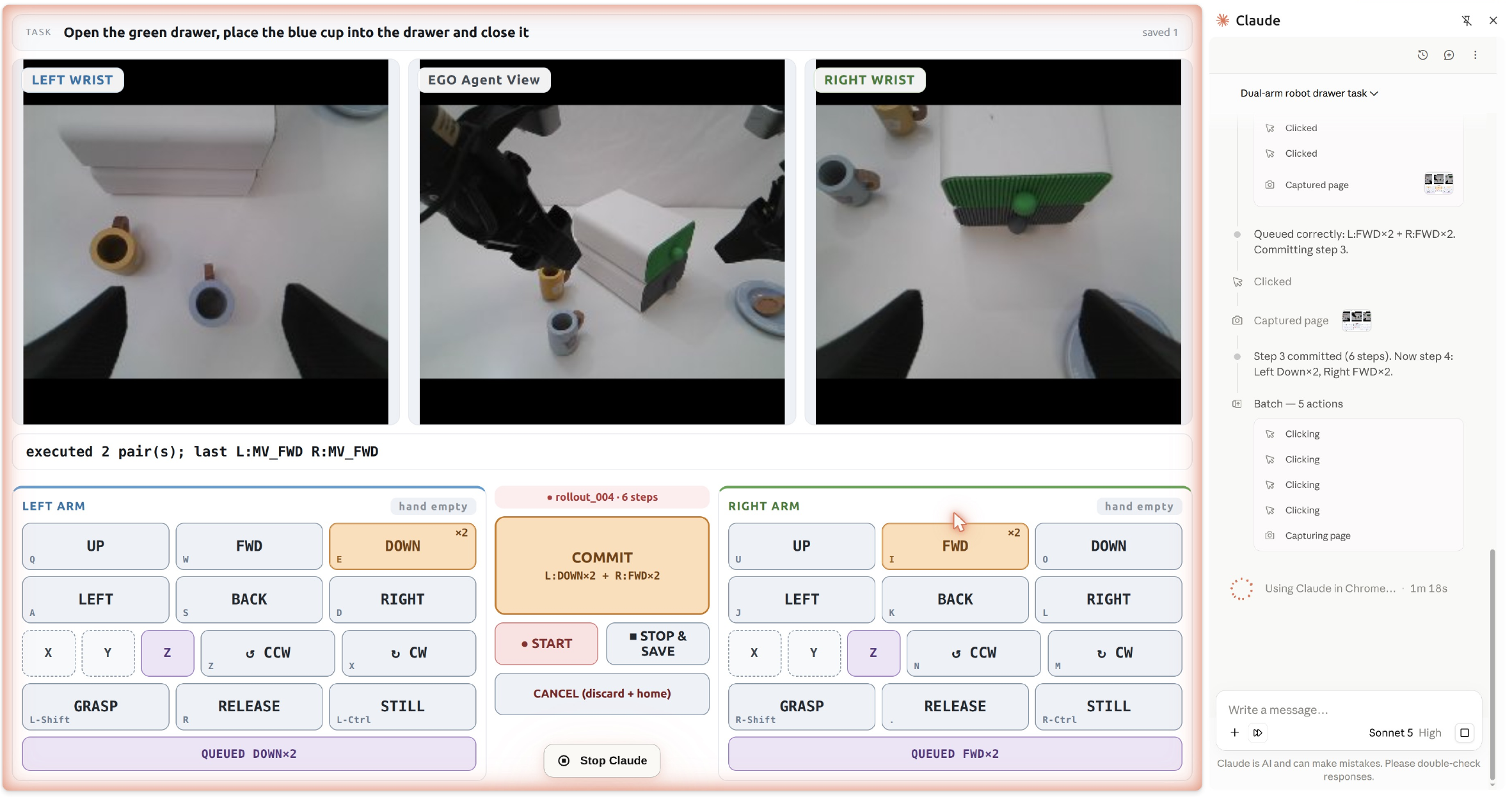}
\caption{\textbf{The GUMI interface.} GUMI is a GUI-based manipulation interface that enables humans and frontier agents to autonomously collect demonstrations through the same semantic controls.}
  \label{fig:gumi}
\end{figure}

\paragraph{A shared interface for humans, agents, and policy learning}
Because the semantic action space in Sec.~\ref{sec:interface} is discrete and directly operable, it can be naturally exposed through a lightweight graphical interface. Building on this property, we develop \textbf{GUMI}, a GUI-based manipulation interface that allows humans and agents to operate robots using the same semantic action units.
As shown in Fig.~\ref{fig:gumi}, each unit maps to a labeled control and keystroke: humans can ``play'' the robot from the keyboard, computer-use agents can operate the same GUI, and general VLM agents can predict the units directly.
At each step, GUMI records the pre-execution observation and selected semantic action, yielding policy-ready pairs $(o_t,a_t)$.
Because each semantic unit is deterministically grounded by the embodiment-specific interpreter, the rollout can also retain corresponding low-level commands and trajectories, allowing one demonstration to train both semantic-action and continuous-control policies.

\paragraph{Flexible and reusable data collection}
GUMI supports step-wise control, queued action chunks, single- and dual-arm operation, and mixed human--agent collection in which humans can intervene to correct agent rollouts.
Unlike conventional teleoperation pipelines that rely on specialized hardware~\citep{zhao2023learning,wu2023gello,chi2024umi} or simulation-specific controls~\citep{zhu2020robosuite,mandlekar2021matters,liu2023libero}, GUMI records demonstrations directly in the shared semantic action space.
The same demonstrations can therefore be reused across embodiments whose interpreters implement the same units, while the same workflow applies in simulation and the real world without specialized teleoperation hardware.
Its lightweight, digitally accessible design further supports remote data collection without requiring physical colocation with the robot.

\section{Experiments}
\label{sec:experiments}

\subsection{Experimental Setup}
\label{sec:exp_setup}

\begin{wrapfigure}{r}{0.62\linewidth}
  \centering
  \vspace{-5.0\baselineskip}
  \begin{subfigure}[b]{0.495\linewidth}
    \centering
    \includegraphics[width=\linewidth]{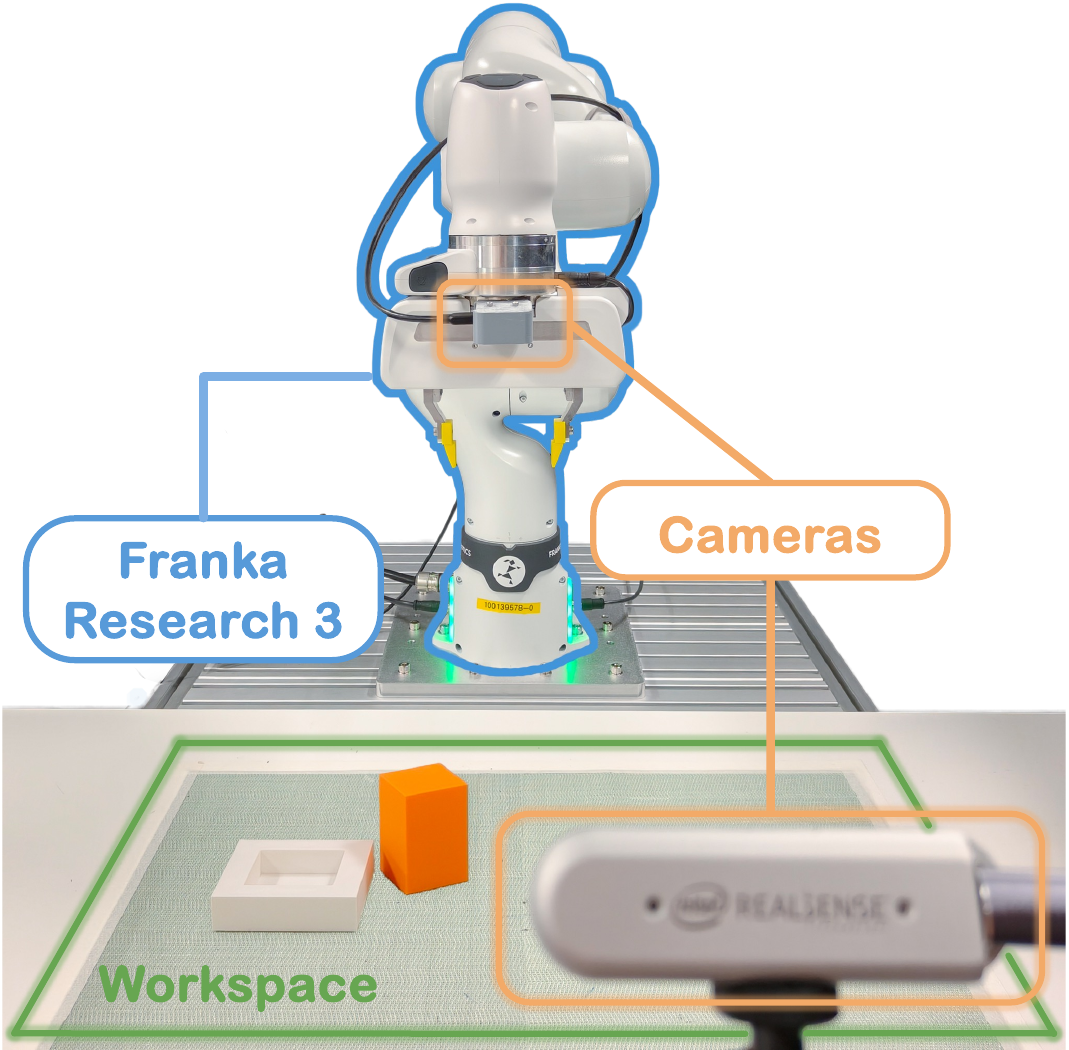}
    \caption{Franka}
    \label{fig:setup_franka}
  \end{subfigure}
  \hfill
  \begin{subfigure}[b]{0.495\linewidth}
    \centering
    \includegraphics[width=\linewidth]{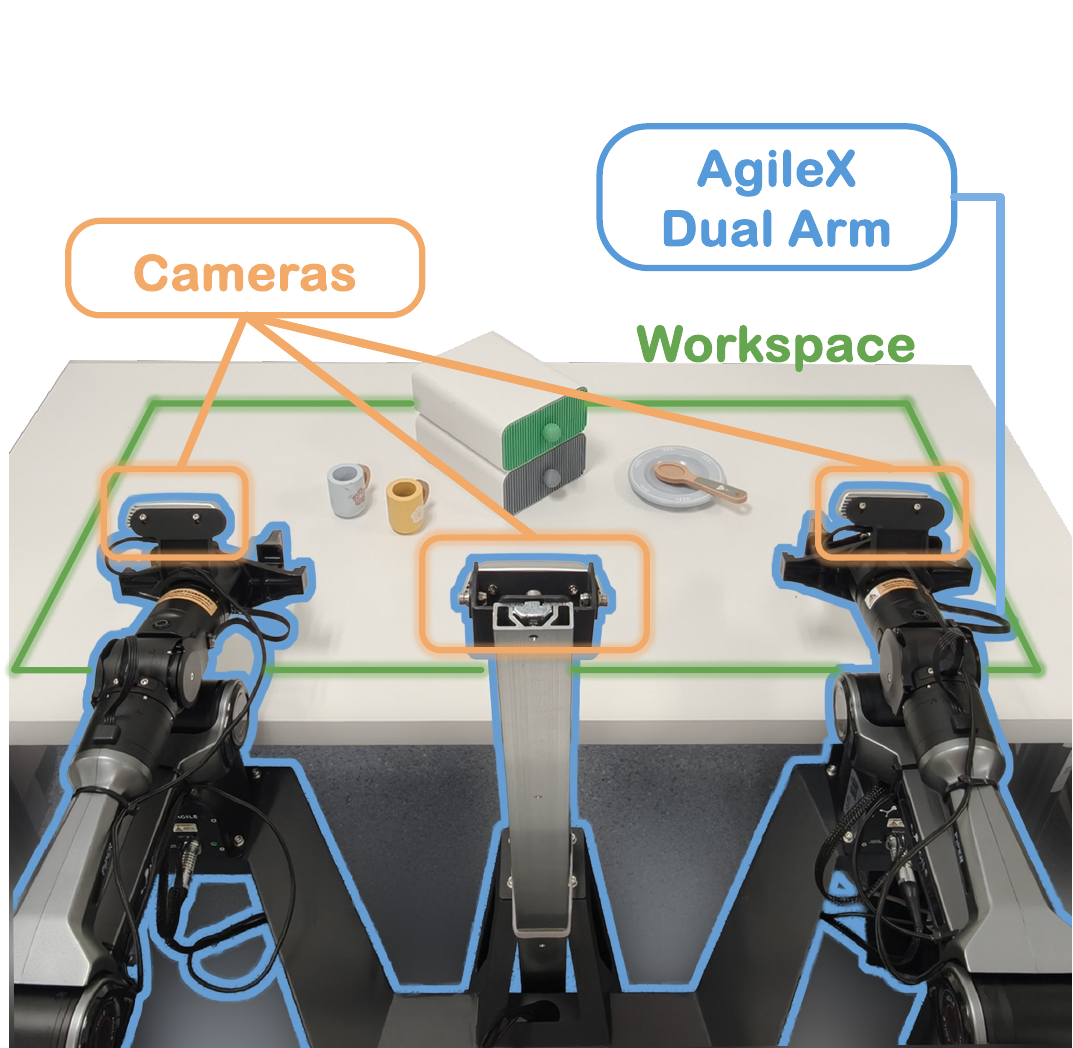}
    \caption{AgileX (dual-arm)}
    \label{fig:setup_dual}
  \end{subfigure}
  \vspace{-1.2\baselineskip}
  \caption{Real-robot rigs.}
  \label{fig:setup}
  \vspace{-1.0\baselineskip}
\end{wrapfigure}

\paragraph{Hardware}
Our experiments use two robot platforms, each equipped with parallel-jaw grippers (Figure~\ref{fig:setup}). The first is a 7-DoF Franka Research 3 arm, observed by an exocentric Intel RealSense D435 facing the workspace and a wrist-mounted Intel RealSense D405, providing two views. The second is a bimanual AgileX rig with two 6-DoF arms, observed by three Orbbec Dabai DC1 cameras: an egocentric view shared by both arms and one wrist view per wrist. Local models are served on a single RTX~5090 GPU, while frontier models are accessed through their
APIs.

\paragraph{Tasks and metrics}
We construct ten real-robot manipulation tasks by pairing five objects with two target receptacles (a plate and a bowl). Each task requires locating, approaching, grasping, transporting, and placing appropriately. Objects span diverse physical properties: rigid geometry (block), irregular shape (banana), rolling dynamics (tennis ball), deformability (teddy bear), and precision manipulation (chess piece). For fine-tuned open-sourced VLMs, only the block, banana, and tennis ball appear in the demonstrations. The teddy bear and chess piece are held out for OOD evaluation.
Beyond these tasks, Secs.~\ref{sec:analysis} and~\ref{sec:ablation} introduce targeted scenarios to probe broader capabilities and design choices.
We report success rate and average steps per episode. Unless noted otherwise, we run 10 trials per task with randomized object placements. Episodes are capped at 50 steps, with timeouts counted as failures.

\paragraph{VLM agents and harness} 
(1) \textbf{Frontier VLM as zero-shot agents (ZS mode).}
Unless otherwise specified, we use Gemini-3.1 Pro~\cite{googledeepmind2026gemini31pro} as the default frontier VLM.
We define \emph{thinking effort} as the inference-time reasoning budget allocated to the VLM, and consider three levels---\textit{low}, \textit{medium}, and \textit{high}---with \textit{medium} used by default; other VLM backbones and reasoning budgets are analyzed in our ablations.
The Adaptive Step plugin uses two translation step sizes: a 2\,cm fine step when the target is visible in the wrist view and a 4\,cm coarse step otherwise.
This simple configuration suffices for basic pick-and-place tasks, while the VLM can flexibly adapt to finer step granularities without fine-tuning (Sec.~\ref{sec:analysis}).
Action history retains the five most recent actions.
All plugins in Sec.~\ref{sec:harness} are enabled by default, except Situated Planning and Visual Prompt, which are activated only when needed.
(2) \textbf{Fine-tuning small VLMs (FT mode).}
We use Qwen3.5-2B~\cite{qwen3.5} by default and analyze different model capacities in our ablations.
We apply rank-$64$ LoRA~\cite{hu2021lora} adapters to all language-model linear layers while freezing the vision encoder and multimodal projector, updating only about $3\%$ of parameters.

\paragraph{Baselines}
We compare against three representative families of robot-control systems: (1) \textbf{VLA} baselines directly map observations and instructions to low-level robot actions, including $\pi_{0.5}$~\citep{black2025pi_05} and GR00T~\citep{bjorck2025gr00t}. For a controlled comparison, both are fine-tuned on continuous end-effector trajectories converted from the same demonstrations used to train our VLM policy. (2) \textbf{VLA-centric agents} retain a VLA as the main physical executor while adding an outer agentic layer for planning, monitoring, or intervention, represented by Harness VLA~\citep{zhang2026harness} (H-VLA in tables) and Goal-VLA~\citep{chen2025goal} (G-VLA in tables). (3) \textbf{Code-as-policy agents} translate high-level reasoning into executable programs or API calls, represented by CaP-X~\citep{fu2026cap} and RATS~\citep{zhang2026playful}.

\paragraph{Demonstration data collection}
For open-source VLM training and trainable VLA baselines, we collect demonstrations through GUMI using human keyboard control and frontier-VLM rollouts via the browser interface. 
In total, we collect $164$ real-robot episodes ($7.8$K decision steps) under a shared $2$\,cm translation: $101$ episodes on the 7-DoF Franka ($5.0$K steps), $63$ episodes on the single-arm AgileX ($2.8$K steps). 
Our fine-tuned policy and compared VLA baselines jointly train on demonstrations from both embodiments. 
For sim-to-real experiments, all compared models are trained on $230$ simulated episodes ($13.5$K steps) collected through the same interface, spanning two simulators: $100$ episodes in ManiSkill~\cite{tao2024maniskill3} with the stock Panda hand, and $130$ episodes over 12 pick-and-place tasks in RoboLab~\cite{yang2026robolab}. More details can be found in the Appendix~\ref{appendix:data}.

\subsection{Main Results: Generalization across Task, Environment, and Embodiments}
\label{sec:main_results}

\begin{table}[t!]
  \centering
  \caption{Performance across three levels of generalization on real robots. ZS and FT denote Show-Harness in zero-shot mode (Gemini-3.1 Pro with medium thinking effort) and fine-tuned mode (Qwen3.5-2B).}
  \label{tab:main}
  \small
  \setlength{\tabcolsep}{4pt}
  \newcolumntype{Y}{>{\centering\arraybackslash}X}
  \begin{NiceTabularX}{\linewidth}{l *{8}{Y}}
  \CodeBefore
    \begin{tikzpicture}
      \fill[slBlue!15]   ([yshift=-\defaultaddspace]row-3-|col-1)  rectangle (row-4-|col-10);
      \fill[slYellow!15] ([yshift=-\defaultaddspace]row-15-|col-1) rectangle (row-16-|col-10);
      \fill[slRed!15]    ([yshift=-\defaultaddspace]row-22-|col-1) rectangle (row-23-|col-10);
    \end{tikzpicture}
  \Body
    \toprule
    \Block{2-1}{Setting} & \Block{1-2}{VLA} & & \Block{1-2}{VLA-centric agent} & &
      \Block{1-2}{\shortstack[c]{Code-as-policy agent}} & & \Block{1-2}{\textbf{\MODEL}} & \\
    \cmidrule(lr){2-3} \cmidrule(lr){4-5} \cmidrule(lr){6-7} \cmidrule(lr){8-9}
    & \mbox{$\pi_{0.5}${\scriptsize}}
    & \mbox{GR00T{\scriptsize}}
    & \mbox{H-VLA{\scriptsize}}
    & \mbox{G-VLA{\scriptsize}}
    & \mbox{CaP-X{\scriptsize}}
    & \mbox{RATS{\scriptsize}}
    & ZS & FT \\
    \midrule
    \addlinespace
    \Block[l]{1-*}{\textbf{Cross-Task}\quad\emph{10 object--receptacle tasks; $\dag$: unseen in fine-tuning; 10 trials / task}} \\
    Block $\to$ Plate        & 6 / 10 & 5 / 10 & 7 / 10 & 2 / 10 & 5 / 10 & 6 / 10 & 10 / 10 & 10 / 10 \\
    Banana $\to$ Plate       & 7 / 10 & 7 / 10 & 8 / 10 & 3 / 10 & 8 / 10 & 8 / 10 & 10 / 10 & 10 / 10 \\
    Tennis $\to$ Plate       & 3 / 10 & 3 / 10 & 3 / 10 & 1 / 10 & 4 / 10 & 6 / 10 & 8 / 10 & 8 / 10 \\
    Teddy$^\dag$ $\to$ Plate   & 5 / 10 & 4 / 10 & 5 / 10 & 1 / 10 & 4 / 10 & 6 / 10 & 10 / 10 & 8 / 10 \\
    Chess$^\dag$ $\to$ Plate & 1 / 10 & 0 / 10 & 3 / 10 & 0 / 10 & 2 / 10 & 4 / 10 & 10 / 10 & 9 / 10 \\
    Block $\to$ Bowl         & 5 / 10 & 4 / 10 & 6 / 10 & 2 / 10 & 5 / 10 & 5 / 10 & 10 / 10 & 10 / 10  \\
    Banana $\to$ Bowl        & 6 / 10 & 6 / 10 & 7 / 10 & 2 / 10 & 7 / 10 & 8 / 10 & 6 / 10 & 7 / 10 \\
    Tennis $\to$ Bowl        & 2 / 10 & 2 / 10 & 3 / 10 & 1 / 10 & 3 / 10 & 5 / 10 & 7 / 10 & 8 / 10 \\
    Teddy$^\dag$ $\to$ Bowl    & 3 / 10 & 4 / 10 & 5 / 10 & 1 / 10 & 4 / 10 & 5 / 10 & 8 / 10 & 7 / 10 \\
    Chess$^\dag$ $\to$ Bowl  & 1 / 10 & 0 / 10 & 3 / 10 & 0 / 10 & 2 / 10 & 4 / 10 & 10 / 10 & 9 / 10 \\
    \rowcolor{slBlue!5}
    {Average (\%)}     & 39.0 & 35.0 & 50.0 & 13.0 & 44.0 & 57.0 & 89.0 & 86.0 \\
    \addlinespace
    \Block[l]{1-*}{\textbf{Cross-Environment}\quad\emph{2 tasks (Block or Banana $\to$ Plate); 10 trials / task}} \\
    Background               & 11 / 20 & 10 / 20 & 15 / 20 & 4 / 20 & 12 / 20 & 14 / 20 & 20 / 20 & 18 / 20 \\
    Lighting                 & 9 / 20  & 9 / 20  & 14 / 20 & 5 / 20 & 11 / 20 & 13 / 20 & 20 / 20 & 19 / 20 \\
    Viewpoint                & 10 / 20 & 7 / 20  & 10 / 20 & 2 / 20 & 10 / 20 & 12 / 20 & 20 / 20 & 19 / 20 \\
    Distractors              & 10 / 20 & 8 / 20  & 12 / 20 & 1 / 20 & 9 / 20  & 13 / 20 & 20 / 20 & 19 / 20 \\
    Sim-to-real              & 0 / 20  & 0 / 20  & --    & --   & --    & --    & --      & 13 / 20 \\
    \rowcolor{slYellow!5}
    {Average (\%)}      & 40.0 & 34.0 & 63.8 & 15.0 & 52.5 & 65.0 & 100.0 & 88.0 \\
    \addlinespace
    \Block[l]{1-*}{\textbf{Cross-Embodiment}\quad\emph{5 Plate tasks; 10 trials / task on each arm}} \\
    Franka (7-DoF)           & 22 / 50 & 19 / 50 & 26 / 50 & 7 / 50 & 23 / 50 & 30 / 50 & 48 / 50 & 45 / 50 \\
    AgileX (6-DoF)           & 19 / 50 & 17 / 50 & 23 / 50 & 4 / 50 & 20 / 50 & 22 / 50 & 45 / 50 & 42 / 50 \\
    \rowcolor{slRed!5}
    {Average (\%)}      & 41.0 & 36.0 & 49.0 & 11.0 & 43.0 & 52.0 & 93.0 & 87.0 \\
    \bottomrule
    \CodeAfter
      \begin{tikzpicture}[slGreen!50, line width=0.8pt, rounded corners=2pt]
        \draw ([yshift=-\defaultaddspace]row-3-|col-8)
            -- ([yshift=-1.5pt]row-1-|col-8)
            -- ([yshift=-1.5pt]row-1-|col-10)
            -- ([yshift=1.5pt]row-26-|col-10)
            -- ([yshift=1.5pt]row-26-|col-8)
            -- (row-23-|col-8);
        \draw (row-4-|col-8)  -- ([yshift=-\defaultaddspace]row-15-|col-8);
        \draw (row-16-|col-8) -- ([yshift=-\defaultaddspace]row-22-|col-8);
      \end{tikzpicture}
  \end{NiceTabularX}
\end{table}

We evaluate three levels of generalization as summarized in Tab.~\ref{tab:main}. Cross-task covers all ten object--receptacle tasks. Cross-environment evaluates background, lighting, viewpoint, distractor, and sim-to-real shifts not seen during training or the standard test setting. For sim-to-real, trainable methods use only simulated demonstrations collected through the same GUMI interface and are evaluated on the real Franka. Cross-embodiment evaluates transfer between the Franka and AgileX arms: the zero-shot agent switches embodiments solely through the embodiment-specific interpreter, while the fine-tuned policy is co-trained on demonstrations from both embodiments and evaluated directly on each arm.

Table~\ref{tab:main} shows that both ZS (zero-shot frontier models) and FT (fine-tuned small open-source models) consistently outperform representative baselines across task, environment, and embodiment shifts. The advantage persists on held-out object combinations and extends to sim-to-real transfer, where FT succeeds using only simulated demonstrations while trainable VLA baselines fail. Cross-embodiment results further show that the same semantic interface transfers effectively between Franka and AgileX. Together, these results suggest that Show-Harness provides a scalable interface spanning frontier zero-shot agents and small fine-tuned models.

\subsection{Capability Analysis: Physical and Semantic Adaptability}
\label{sec:analysis}

\begin{figure}[t]
  \centering
  \includegraphics[width=\linewidth]{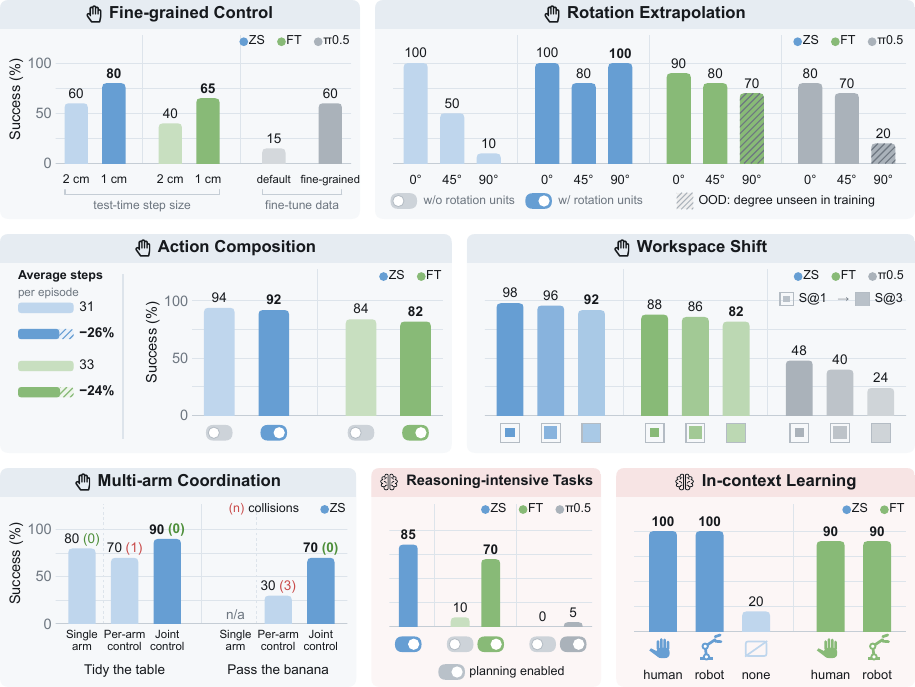}
 \caption{Capability analysis of Show-Harness across physical (\raisebox{-0.35ex}{\includegraphics[height=1.9ex]{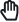}}) and semantic (\raisebox{-0.45ex}{\includegraphics[height=2.1ex]{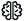}}) adaptability.}
  \label{fig:generalization}
\end{figure}

Sec.~\ref{sec:main_results} establishes broad generalization across task, environment, and embodiment shifts.
We next probe a harder question: \textbf{how well can Show-Harness enable foundation models to adapt to new physical and semantic demands?}
Through targeted scenarios, we study two complementary forms of adaptability: \emph{physical adaptability}, which accommodates changes in motion precision, composition, workspace, and embodiment without policy retraining, and \emph{semantic adaptability}, which handles tasks requiring reasoning or in-context learning from novel demonstrations (as shown in Fig.~\ref{fig:generalization}).

\subsubsection{Physical Adaptability}

\textbf{Fine-grained control.}
We evaluate block stacking and peg insertion, which require finer motion than standard pick-and-place.
We simply reduce the interpreter step size from 2\,cm to 1\,cm, without changing the VLM--action interface or retraining either model.
This improves ZS from 60\% to 82\% and FT from 40\% to 65\%.
In contrast, $\pi_{0.5}$ achieves only 18\% with the same demonstrations as FT, reaching 62\% only after additional fine-grained training.
This highlights the benefit of separating semantic decisions from metric execution: Show-Harness-equipped VLM agents adapt to new precision requirements through the interpreter alone, without task-specific relearning.

\textbf{Action composition.}
On the five Plate tasks, composing two orthogonal translation units into one diagonal displacement substantially reduces execution steps without a notable drop in success.
This shows that new action compositions can be introduced through the interpreter without retraining.

\textbf{Rotation extrapolation.}
We test compositional extrapolation on a carrot-grasping task with the carrot oriented at 0$^\circ$, 45$^\circ$, or an unseen 90$^\circ$ relative to the gripper, while each rotation unit changes orientation by 15$^\circ$.
As shown in Fig.~\ref{fig:generalization}, ZS remains robust across all angles, while FT reaches 70\% at the unseen 90$^\circ$ orientation despite training only on 0$^\circ$ and 45$^\circ$ demonstrations; $\pi_{0.5}$ reaches 20\%.
This shows that incremental rotation enables larger unseen orientation changes through repeated unit composition, whereas continuous action regression remains more tied to the demonstrated angle range.

\textbf{Workspace shift.}
On the five Plate tasks, we evaluate nested workspace regions following~\cite{chen2026escaping}, from the core 25\% (S@1) to 90\% near the boundary (S@3).
Show-Harness degrades only mildly as the workspace expands, whereas $\pi_{0.5}$ drops sharply.
This suggests that visually grounded semantic decisions make Show-Harness-equipped VLM agents less sensitive to workspace shifts, while $\pi_{0.5}$ remains more tied to the training distribution.

\textbf{Multi-arm coordination.}
We evaluate two 20-trial tasks on an AgileX pair: \emph{tidy the table}, where each arm clears nearby objects, and \emph{pass the banana}, which requires handover before placement.
We compare two independent single-arm agents with a joint policy that predicts both arms' actions at each step.
Joint decision making substantially improves success on both tasks while eliminating collisions, showing that Show-Harness enables explicit inter-arm coordination through joint action prediction.

\subsubsection{Semantic Adaptability}

\textbf{Reasoning-intensive tasks.}
We evaluate two tasks requiring high-level reasoning before manipulation: locating a block hidden under one of three cups and arranging scattered letters into ``SHOW''.
ZS with Situated Planning achieves 85\%, while FT and $\pi_{0.5}$ alone reach only 10\% and 0\%.
With the same Gemini-generated subtask instructions, FT rises to 70\%, whereas $\pi_{0.5}$ remains at 5\%.
This suggests that Show-Harness better preserves the instruction-following flexibility of VLMs by grounding novel instructions through a reusable semantic action space.

\textbf{In-context learning from video.}
We ask the policy to put away three objects in the order shown by a human or robot demonstration.
Without a demonstration, ZS receives only the instruction \emph{tidy up} and succeeds in just 20\%, since the required order is unspecified.
With a video demonstration, ZS follows the demonstrated order and succeeds in 20/20 trials from either source, while FT reaches 95\% when conditioned on the task outline extracted by the same planner.
This shows that Show-Harness enables strong visual in-context learning for robot control.

\subsection{Ablation Studies}
\label{sec:ablation}

\begin{figure}[htb]
  \centering
  \includegraphics[width=\linewidth]{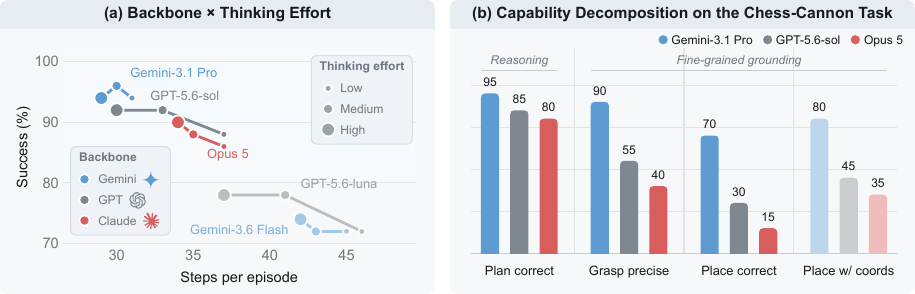}
  \caption{Effect of frontier VLM choice and thinking effort.}


  \label{fig:breadth}
\end{figure}

\subsubsection{Frontier VLM Choice and Thinking Effort}
We evaluate different frontier VLMs on the five Plate tasks (Fig.~\ref{fig:breadth} (a)), and further probe the strongest three models on a chess-cannon placement task (20 trails) that separates planning from fine-grained grounding (Fig.~\ref{fig:breadth} (a)).
As shown in Fig.~\ref{fig:breadth} (a), zero-shot performance generally improves with stronger frontier VLMs, broadly tracking their underlying model capability, while increasing thinking effort mainly reduces redundant interaction steps with little gain in success and can incur higher wall-clock cost (e.g., $3.4\times$ for GPT-5.6-sol). We also find that all models follow instructions reliably, with over 98\% of responses producing valid action units.
Fig.~\ref{fig:breadth} (b) shows that planning is also not the main bottleneck; errors concentrate on fine-grained grasping and placement. Providing target bounding boxes further improves performance, highlighting the value of explicit visual grounding cues.


\subsubsection{Fine-Tuned Backbone Scaling}

\begin{figure}[htp]
  \centering
  \includegraphics[width=\linewidth]{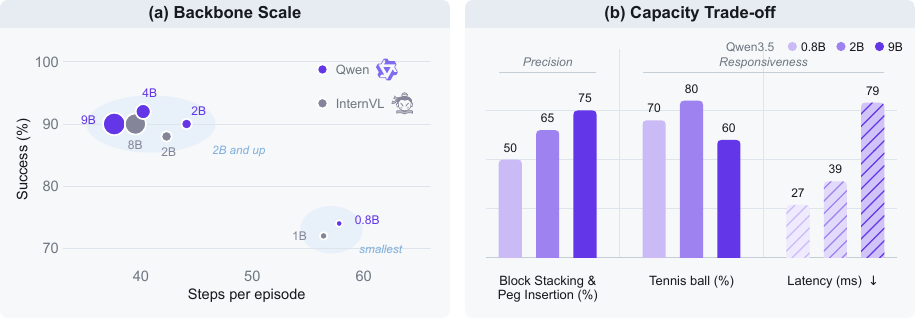}
  \caption{ 
  (a) Effect of fine-tuned backbone scaling over the five Plate tasks; bubble area is proportional to parameter count.
  (b) Three Qwen3.5 capacities across fine-grained tasks.}
  \label{fig:scale}
\end{figure}

As shown in Fig.~\ref{fig:scale}, performance is already strong at 2B, while larger backbones mainly help fine-grained tasks such as stacking and peg insertion. The 1B-level models instead suffer from excessive local adjustments near the target, leading to longer episodes. Smaller models can nevertheless outperform larger ones on the tennis-ball task, where timely correction of moving objects matters more than fine precision. Overall, the 2B backbone offers a good balance between precision and responsiveness.

\subsubsection{Harness Plugins}

\begin{figure}[t]
  \centering
  \includegraphics[width=\linewidth]{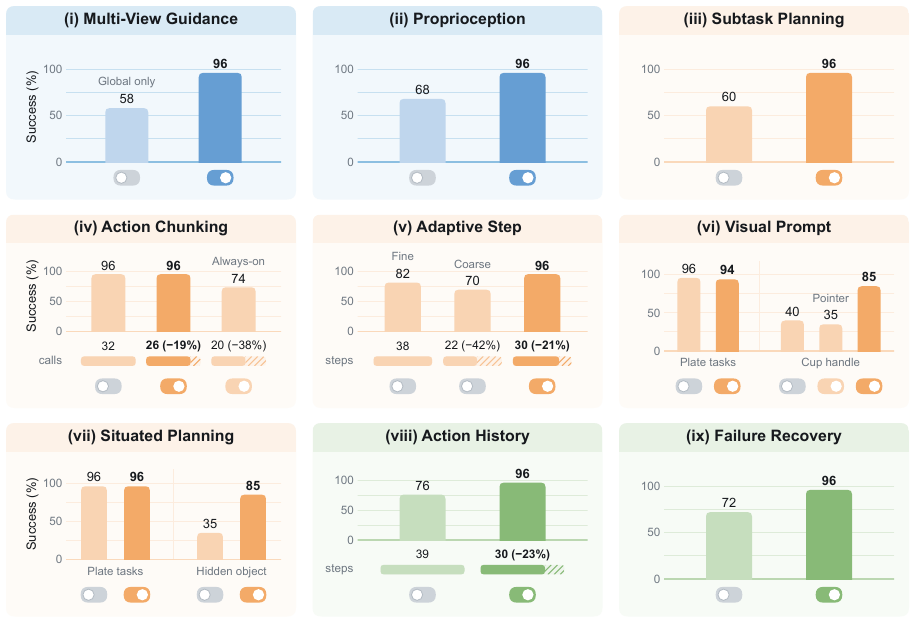}
  \caption{Plugin ablations on the real Franka arm using Gemini-3.1 Pro as a zero-shot agent.}
  \label{fig:plugins}
\end{figure}

Fig.~\ref{fig:plugins} evaluates the harness plugins on the real Franka arm with the zero-shot agent (Gemini-3.1 Pro).
Default plugins follow a leave-one-out protocol on the five Plate tasks.
Visual Prompt and Situated Planning are instead added to the default configuration and evaluated on dedicated scenarios requiring handle-aware grasping and hidden-object search under three inverted cups (20 trails each), respectively.

\noindent \textbf{Multi-View Guidance} substantially improves performance over global-view input alone, particularly for small objects requiring precise alignment. This is because multiple views enhance spatial perception, while the wrist view provides especially useful cues for fine-grained discrimination.

\noindent \textbf{Proprioception}. Removing proprioceptive state substantially degrades performance, particularly on objects where visual cues alone are ambiguous. Compact state signals such as gripper height and contact provide reliable guidance when appearance or depth makes visual estimation uncertain.

\noindent \textbf{Subtask Planning}. 
Without planning, success drops to 60\%, with the model often dragging objects toward the plate without lifting. Subtask decomposition makes the intended manipulation sequence explicit through locally valid stages and visually checkable completion criteria.

\noindent \textbf{Action Chunking}. 
Disabling chunking preserves 96\% success but increases model calls, whereas forcing it throughout reduces calls but drops success to 74\%.
The result favors selective chunking: compress redundant transport steps while preserving fine-grained feedback near interaction.

\noindent \textbf{Adaptive Step}. 
Fine-only control is precise but inefficient and prone to timeouts, while coarse-only control is faster but prone to overshoot. Adaptive switching based on target visibility balances both, achieving 96\% success with 30 steps per episode on average.

\noindent \textbf{Visual Prompt}.
The plugin has little effect on regular tasks, supporting its default-off design, but substantially improves handle-aware grasping (40\% to 85\%).
The gain comes not from the visual marker alone, but from explicitly aligning the language instruction with the marked interaction point.

\noindent \textbf{Situated Planning}.
The plugin has no effect on regular tasks, but substantially improves hidden-object search (35\% to 85\%).
It defers unresolved decisions until sufficient visual evidence is available, enabling targeted exploration instead of premature planning.


\noindent \textbf{Action History}.
Removing action history reduces success and increases timeouts, often due to oscillation between opposing actions.
Recording recent actions exposes such loops and prevents repeated reversals, providing lightweight memory for more stable closed-loop control.

\noindent \textbf{Failure Recovery}.
Removing recovery reduces success to 72\%, with the largest drops on hard-to-grasp objects. The main failure mode is an undetected empty grasp; empty-grasp detection triggers a retry, preventing the agent from continuing with an empty gripper.

\subsubsection{Action-space Representation}
We study what makes the semantic action space effective through a controlled ablation with Gemini-3.1 Pro, varying only the representation of the six translational action units while keeping the remaining harness fixed.
A $2\times2$ design compares: (A) semantic action names with written conventions specifying their physical effects (our default), (B) semantic names only, (C) arbitrary symbols with conventions, and (D) arbitrary symbols only.
For (D), which provides no prior action semantics, the agent probes unknown symbols, infers their effects from before/after observations, and records the mappings for later use.
Each variant is evaluated on the Block/Banana~$\to$~Plate tasks over 20 trials in total.

\begin{figure}[t]
  \centering
  \includegraphics[width=\linewidth]{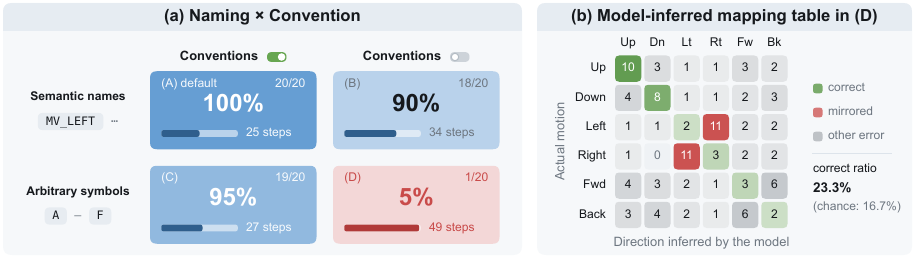}
  \caption{(a) Ablation on action-space representation.
  (b) Action mappings inferred under setting (D) over 20 episodes, where the model probes unknown symbols and infers their effects from observation changes.}
  \label{fig:action_repr}
\end{figure}

Fig.~\ref{fig:action_repr} (a) shows that arbitrary symbols with explicit conventions (C) nearly match the default, while semantic names alone (B) remain usable but less efficient.
This indicates that conventions provide most of the grounding, with semantic names serving mainly as a useful prior.
Setting (D) succeeds in only 1/20 episodes with only 23.3\% of inferred mappings correct (Fig.~\ref{fig:action_repr} (b)).
Explicit conventions therefore avoid the ambiguity of inferring physical action effects from visual changes alone.

\subsection{Qualitative Analysis}

\begin{figure}[t!]
  \centering
  \includegraphics[width=\linewidth]{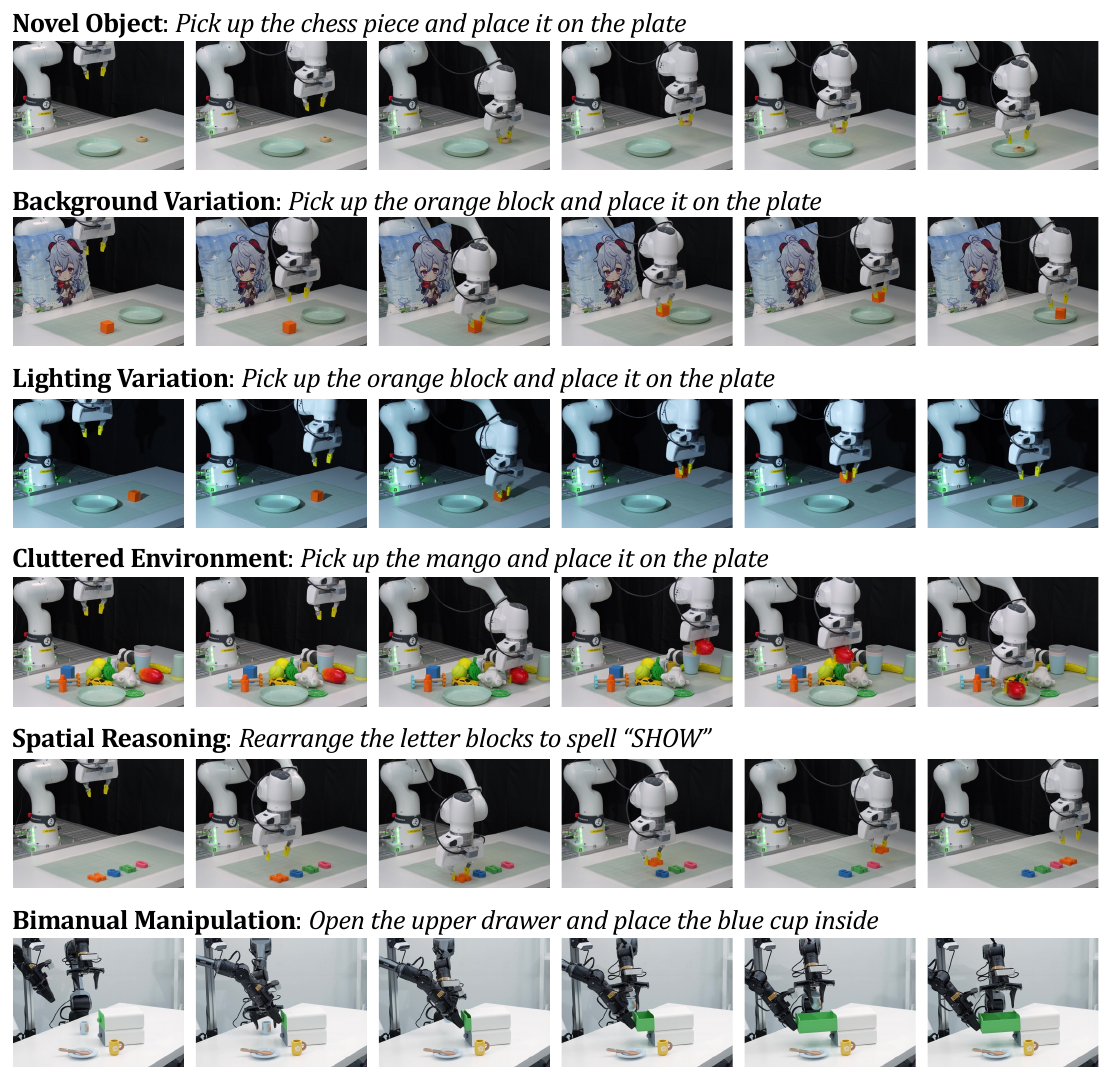}
  \caption{
Qualitative real-world demonstrations of Show-Harness across diverse tasks and conditions.
}
  \label{fig:real_demos}
\end{figure}

We further provide qualitative results on a diverse set of real-world manipulation scenarios. As shown in Fig.~\ref{fig:real_demos}, Show-Harness successfully handles a broad range of visual and physical variations, including novel objects, background changes, lighting changes, cluttered scenes, spatial-reasoning tasks, and bimanual control.
Beyond standard pick-and-place, the same interface supports more challenging behaviors such as rearranging letter blocks to satisfy a semantic goal, and coordinating two arms to open a drawer and place an object inside.
These examples qualitatively demonstrate that the proposed semantic action interface can be reused across substantially different task structures and manipulation requirements without redesigning the model-facing action space.
We further provides qualitative comparison with $\pi_{0.5}$ at Appendix~\ref{appendix:compare_pi}.

\section{Conclusion and Limitations}

In this work, we introduced \textbf{Show-Harness}, an \emph{Embodied Harness} that unlocks embodied capability from foundation VLMs through a compact semantic action interface.
Show-Harness exposes an action space that is semantic enough for VLMs to naturally reason over yet fine-grained enough for direct physical control, keeping the VLM engaged in stepwise physical decisions within a closed perception--reasoning--action--feedback loop.
Building on this shared interface, we further introduced \textbf{GUMI}, which enables humans and agents to collect robot demonstrations through a GUI without specialized teleoperation hardware.
Extensive experiments demonstrate strong generalization and adaptability of Show-Harness-enabled VLM agents, suggesting that the right interface can unlock substantial embodied capability already present in foundation VLMs, with minimal embodiment-specific adaptation.

Despite its effectiveness, Show-Harness is currently evaluated primarily on single- and dual-arm manipulation with parallel-jaw grippers. Extending the framework to more complex embodiments (e.g., humanoids or dexterous hands) is a valuable direction. Future work may also enrich the perception side with additional embodied modalities, such as tactile and force feedback, to support more contact-rich and fine-grained physical interaction.

\bibliographystyle{plainnat}   
\bibliography{A-reference}

\clearpage
\section{Appendix}

\subsection{VLM Fine-Tuning Details}
\label{sec:impl}

For lightweight VLM fine-tuning, we train for $40$ epochs on the 7.9K single-arm samples at a learning rate of $1\times10^{-4}$ under a cosine schedule with a warmup ratio of $0.1$, using bf16, $256\times256$ views, and an effective batch size of $32$.
The fine-tuning is lightweight and can be performed on 24\,GB-class GPUs.
In our experiments, fine-tuning the Qwen3.5-2B model takes less than 2 hours on a single H200.

\begin{table}[htb]
  \centering
  \caption{Demonstration corpus collected through GUMI.}
  \label{tab:demodata}
  \small
  \setlength{\tabcolsep}{4pt}
  \newcolumntype{C}{>{\centering\arraybackslash}X}
  \begin{tabularx}{\linewidth}{@{}llCCCCC@{}}
    \toprule
    & Platform & Tasks & Episodes & Ep./Task & Steps & Steps/Ep. \\
    \midrule
    \multirow{3}{*}{Real Robots}
      & Franka (7-DoF)   &  9 & 101 &  11.2 & 4969 & 49.2 \\
      & AgileX (6-DoF)   & 10 &  63 &   6.3 & 2805 & 44.5 \\
    \cmidrule(l){2-7}
      & \textit{total}   & 19 & 164 & 8.6 & 7774 & 47.4 \\
    \midrule
    \multirow{3}{*}{Simulation}
      & ManiSkill        &  1 & 100 & 100.0 &  5840 & 58.4 \\
      & RoboLab          & 12 & 130 &  10.8 &  7683 & 59.1 \\
    \cmidrule(l){2-7}
      & \textit{total}   & 13 & 230 & 17.7 & 13523 & 58.8 \\
    \bottomrule
  \end{tabularx}
\end{table}


\subsection{Demonstration Data}
\label{appendix:data}
Table~\ref{tab:demodata} breaks the training corpus down per platform. The corpus is released at \url{https://huggingface.co/showlab/Show-Harness-Data}.

The real-robot demonstration covers $19$ tasks, nine distinct object--target combinations on the Franka and ten on the AgileX, concentrated on pick-and-place with a few stacking and shelf-rearrangement episodes. 
All real-robot episodes are recorded through GUMI with the same $2$\,cm translation step size, so a semantic unit denotes the same displacement on the Franka and both AgileX configurations.
Each episode stores the third-person and wrist frames for every step together with the emitted unit, the measured end-effector pose, and the gripper width; a training sample pairs the two views at one step with the unit taken at that step, plus one terminal sample per episode that emits \texttt{DONE}. Beyond clean executions, $19$ Franka episodes ($594$ steps) are recorded specifically as grasp recovery: the gripper starts short of, past, or to either side of the intended grasp point, and a close that comes up empty is followed by lifting the gripper and re-approaching rather than transporting an empty hand. These episodes are trimmed so that only the corrective segment is kept, which is why they run about $31$ steps against $53$ for the freely collected ones.

For simulation data, we collect demonstrations from ManiSkill~\cite{tao2024maniskill3} and RoboLab~\cite{yang2026robolab}.
Across both simulators, we keep the model-facing control and observation conventions consistent with the real Franka setup: each semantic translational action unit corresponds to a $2$\,cm single-axis Cartesian translation, and camera views are converted to the same input format used at deployment.
At the same time, we retain substantial variation in rendering, scene composition, and embodiment appearance to preserve a meaningful sim-to-real gap.
ManiSkill provides $100$ tabletop pick-and-place episodes involving an orange block and a coaster, while RoboLab provides $130$ episodes over twelve tasks, including coloured blocks into bowls, left-versus-right bowl selection, canned food and yogurt into containers, a Rubik's cube, and throwing away an apple.

\subsection{Qualitative Comparison}
\label{appendix:compare_pi}

\begin{figure}[t]
  \centering
  \includegraphics[width=\linewidth]{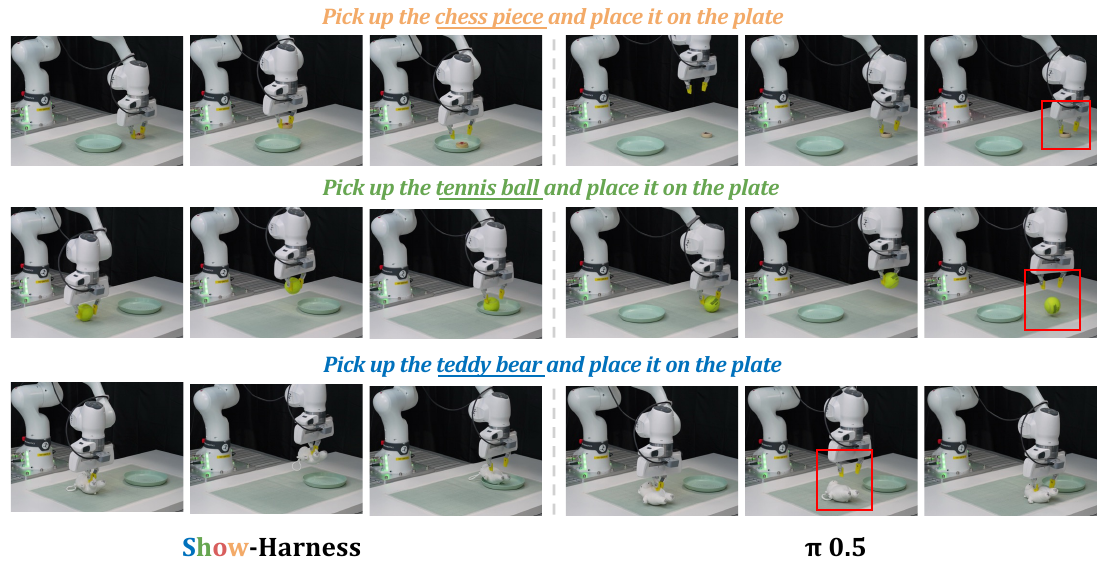}
  \caption{
Qualitative comparison with $\pi_{0.5}$. Show-Harness completes all three examples, while $\pi_{0.5}$ exhibits manipulation failures (e.g., gripper–object collisions and unstable grasps) highlighted by the red boxes.
}
  \label{fig:comparison_pi}
\end{figure}

Fig.~\ref{fig:comparison_pi} provides a controlled qualitative comparison with $\pi_{0.5}$.
For a fair comparison, Show-Harness uses a Qwen3.5-2B VLM fine-tuned on our collected demonstrations, while $\pi_{0.5}$ is fine-tuned on the corresponding low-level action trajectories converted from the same demonstration set.
Under this matched-data setting, Show-Harness consistently completes the manipulation sequence across the chess piece, tennis ball, and teddy bear examples, whereas $\pi_{0.5}$ exhibits grasping or object-interaction failures, as highlighted in red.
These examples qualitatively illustrate the benefit of learning over the shared semantic action interface rather than directly fitting embodiment-specific low-level actions.

\subsection{Runtime Prompts}
\label{app:runtime_prompts}

We report one canonical prompt for each distinct model role rather than enumerating every embodiment- or backend-specific variant.
Braced fields such as \texttt{\{task\}} are populated with the current task, plan state, measurements, or action history immediately before the model call; images are supplied separately through the model API. By default, rotation-related guidance and actions are omitted from the prompt and enabled only for tasks that require end-effector reorientation.

\lstdefinestyle{showharnessprompt}{
  basicstyle=\ttfamily\scriptsize,
  columns=fullflexible,
  keepspaces=true,
  breaklines=true,
  breakatwhitespace=true,
  showstringspaces=false,
  aboveskip=0pt,
  belowskip=0pt,
  tabsize=2
}

\newtcolorbox{showharnesspromptbox}[1]{
  breakable,
  title={#1},
  colback=white,
  colframe=gray,
  coltext=black,
  boxrule=0.8pt,
  left=1mm,
  right=1mm,
  top=1mm,
  bottom=1mm
}

\subsubsection{Frontier VLM Agent}

Show-Harness wraps the same VLM into two roles within the agentic loop: (1) a planner that decomposes the task into visually verifiable stages, and (2) a controller that selects the next semantic action from the current context to control the robot.
At runtime, the controller template is augmented with the active plugin context and an output contract.

\begin{showharnesspromptbox}{Controller prompt ($\mathcal{P}_{\mathrm{ctrl}}$)}
\begin{lstlisting}[style=showharnessprompt]
TASK: {task}
STAGE: {stage}
TARGET: {target}
AFFORD: {affordance}
Stage goal: {description}
DONE WHEN: {completion}
Gripper now: {gripper_state}
{mem_text}
{recovery}
{proprio}

DIRECTION:
Is STAGE for grasping AND TARGET inside the wrist view?
A) YES -> wrist is the primary guide. Judge AFFORD's position vs the {gripper_color} gripper and take the direction of LARGEST deviation:
- AFFORD to the gripper's left  -> MV_LEFT
- AFFORD to the gripper's right -> MV_RIGHT
- AFFORD near image bottom and far from the grippers -> MV_FWD
- AFFORD between the image top and the grippers -> MV_BACK
- AFFORD roughly centered between the grippers -> MV_DOWN
{rotation}
B) NO -> AgentView is the primary guide. Judge TARGET's position vs the end effector (before GRASP) or holding object (after GRASP), and take the direction of LARGEST deviation:
- TARGET near the image left  -> MV_LEFT
- TARGET near the image right -> MV_RIGHT
- TARGET near the image bottom -> MV_FWD
- TARGET near the image top -> MV_BACK
C) MV_UP when:
- Need to lift the object
- too low to reach TARGET
- retreating after a RELEASE

ATTENTION:
{mem_text_rules}
- DONE only when "DONE WHEN" is already visible in the images

GRIPPER:
- GRASP when BOTH AgentView and Wrist view confirm the {affordance} is clearly between the center of two grippers
- RELEASE only when the held object is above its destination and lowered onto it
{variable_step}
{action_chunk}
Think one visual sentence, then commit.
{output_contract}
\end{lstlisting}
\end{showharnesspromptbox}

The default structured-decoding route fills \texttt{\{output\_contract\}} as follows; reasoning-oriented backbones use the same action alphabet and recover the final unit from their response.

\begin{showharnesspromptbox}{Atomic action output contract}
\begin{lstlisting}[style=showharnessprompt]
Choose exactly one action:
MV_FWD, MV_BACK, MV_LEFT, MV_RIGHT, MV_UP, MV_DOWN, GRASP, RELEASE, DONE
Return JSON only: {"decision":"ONE_ACTION","reasoning":"one visual sentence"}
\end{lstlisting}
\end{showharnesspromptbox}

The default harness adds only short, state-dependent fragments to this controller.
The listing below consolidates the fragments that realize Proprioception, Multi-View Guidance, Action Chunking, Adaptive Step, Action History, and Failure Recovery.
Bracketed labels identify the owning plugin and are not sent to the model.

\begin{showharnesspromptbox}{Default plugin injections ($\Phi_{\mathcal P}$)}
\begin{lstlisting}[style=showharnessprompt]
[proprioception/block]
The gripper is {gap_cm} cm above the table; {step_sizes}. {hint}

[proprioception/step_sizes_fine]
each step moves ~{fine_cm} cm

[proprioception/step_sizes_coarse]
each step moves ~{fine_cm} cm ({coarse_cm} cm)

[proprioception/hint_descend]
If height > {high_cm} cm, MV_DOWN first.

[proprioception/hint_holding]
Holding an object: lift until clear of the table; descend only to place.

[proprioception/descend_stall]
Last MV_DOWN lowered {moved_cm} of {commanded_cm} cm -> already in contact, do NOT MV_DOWN again

[wrist_marker/marker]
WRIST CHECK: begin your reasoning with `WRIST: YES` if the TARGET is visible in the wrist view, else `WRIST: NO`.

[action_chunk/plan]
ACTION PLAN -- only when WRIST: NO (TARGET far): plan your next {step_num} moves as `PLAN: M1, M2, ...` (MV_ tokens only) and set decision to M1. Choose each move from the height and step size so the plan does not overshoot (e.g. never plan more MV_DOWN than the height above the table allows). When WRIST: YES, decide a single move.

[mem_text/recent]
Recent moves, newest first: {moves}

[mem_text/rules]
- If recent moves show GRASP(empty), do not GRASP in place again; prioritize MV_UP, MV_BACK, MV_DOWN, or MV_FWD
- NEVER OSCILLATE: Do NOT choose the opposite of the newest recent move (Pairs: MV_LEFT/MV_RIGHT, MV_FWD/MV_BACK)
- When opposite directions appear in recent moves, prioritize MV_DOWN or MV_UP

[recovery/context]
Recovery: {note}

[recovery/note_empty_grasp]
Empty close; do not retry on an edge/corner. Recenter body and confirm depth.

[recovery/note_unverified_grasp]
Grasp not verified; continue GRASP until width and images show a real hold.

[recovery/note_lost_grasp]
Grasp lost; return to GRASP, recenter the object body, then confirm depth.

[recovery/note_unsettled]
Closed width is unsettled; wait before moving.
\end{lstlisting}
\end{showharnesspromptbox}

\begin{showharnesspromptbox}{Subtask Planning prompt ($\mathcal{P}_{\mathrm{plan}}$)}
\begin{lstlisting}[style=showharnessprompt]
ROLE: SubgoalPlanner
TASK: {task}

{video_ref}

Return an ordered JSON plan:
{{
  "subgoals": [
    {{
      "id": "short_snake_case_id",
      "target": "object or destination",
      "affordance": "visible part or placement region",
      "motion": "semantic stage label",
      "description": "visual strategy for this stage",
      "completion": "visible condition that means this stage is complete"
    }}
  ]
}}

### STRICT RULES

1. Stage Segmentation

-Break the task down into meaningful visual milestones (e.g., GRASP, LIFT, MOVE, PLACE, RELEASE, RETREAT)
-MERGE: Do NOT split immediate pre-grasp steps. Combine approach, align, lower, and close into a single `GRASP` stage
-SEPARATE: Keep lift/clearance after a successful grasp as a separate `LIFT` stage
-RETREAT: After every `RELEASE`, add a `RETREAT` stage that lifts the gripper up

2. Affordance Selection

-ONE specific part -- never alternatives like "left end or middle"
-Must be visible in AgentView & bracketable by open fingers
-Containers/Hollow objects (cups, bowls): Target left/right rim or edge.
-Simple Solid objects (blocks): main body

3. Completion Criteria

-ALL completion conditions MUST be strictly judgeable from raw 2D images
-Movement stages: End with a stable visual spatial relation, NOT a gripper event
-Set-aside placements: If moving an object to another location, ensure it is placed away from the origin point on the table
-MUST distinguish among similar objects

Return JSON ONLY.
\end{lstlisting}
\end{showharnesspromptbox}

The dual-arm deployment instantiates the same structure once per arm and predicts the two actions jointly, adding \texttt{STILL} for an idle arm; we therefore do not repeat the near-isomorphic controller and planner templates.
Likewise, the specialized Situated Planning and Visual Prompt roles are used only in their targeted experiments, with full implementation details provided in our code repository.

\textbf{Video-conditioned in-context planning.}
Video conditioning introduces a distinct input role for learning task procedures from demonstrations.
An analyst converts temporally ordered demonstration frames into a textual task outline inserted at \texttt{\{video\_ref\}} in the planner prompt, allowing the agent to infer the demonstrated procedure in context and reuse it during execution.
Here we show the single-arm version, while the dual-arm variant differs only by attaching an arm identity to each operation.

\begin{showharnesspromptbox}{Reference-video analyst prompt}
\begin{lstlisting}[style=showharnessprompt]
ROLE: DemoVideoAnalyst
You see {num_frames} frames sampled IN ORDER from ONE reference video demonstrating a table-top manipulation performed by ONE arm (a robot gripper, or a human hand standing in for it).
A frame may be a composite of labeled camera panels (front view + wrist view); overlay text is auxiliary -- trust what the imagery shows.

Report exactly what is demonstrated, operation by operation, in the order performed. Capture the details that make replication faithful:
- object: name it so it cannot be confused with similar ones (color/size/position)
- grasp: the exact part grasped (stem, rim, edge, handle, top face, ...)
- destination: where it ends up, with the placement nuance (which side/half, on/into, orientation)
Report only what the frames show; do not invent steps between or beyond them.

Return JSON only:
{{"task":"one line: what the demo achieves","operations":[{{"action":"short verb phrase","object":"...","grasp":"...","destination":"..."}}]}}
\end{lstlisting}
\end{showharnesspromptbox}

\subsubsection{Fine-Tuned Lightweight VLM Agent}

The fine-tuned lightweight VLM directly serves as the controller, without a separate planner by default.
It takes only the task instruction, current multi-view observation, recent action history, and shared action vocabulary as input.

\begin{showharnesspromptbox}{Post-trained VLM prompt ($\mathcal{P}_{\mathrm{min}}$)}
\begin{lstlisting}[style=showharnessprompt]
You are controlling a robot arm with two cameras:
- Agentview: overhead view of the robot and workspace
- Wristview: close-up view from the gripper

Task: {task}
Recent moves, newest first: {recent_moves}

Output exactly one action token:
MV_FWD, MV_BACK, MV_LEFT, MV_RIGHT, MV_UP, MV_DOWN, GRASP, RELEASE, DONE

Check both camera views and choose the next action:
- Use AgentView to locate the target when it is not in wrist view
- Use wrist view to fine-align when target is visible up close
- GRASP when gripper fingers are aligned around the object
- RELEASE when object is above the destination
- DONE when the task is complete: the object is at its destination and the gripper is clear
- Avoid repeating a direction that conflicts with the most recent move

Return the single token only, no punctuation, no explanation:
\end{lstlisting}
\end{showharnesspromptbox}

\end{document}